\documentclass[11pt]{article}
\pdfoutput=1

\usepackage[T1]{fontenc}
\usepackage[utf8]{inputenc}
\usepackage{times}
\usepackage{latexsym}
\usepackage{inconsolata}
\usepackage[final]{ACL2023} 

\usepackage{microtype}

\usepackage{amsmath}
\usepackage{amssymb}
\usepackage{amsthm}

\usepackage{algorithm}
\usepackage{algpseudocode}

\usepackage{booktabs}
\usepackage{multirow}
\usepackage{graphicx} 
\usepackage{graphicx}
\usepackage{tikz}
\usepackage{tabularx} 

\usepackage{xcolor}

\usepackage{hyperref}

\usepackage{natbib}

\usepackage{xcolor}
\usepackage{array} 
\usepackage{tabularx} 
\usepackage{multirow}  
\usepackage{float}  
\usepackage{fancyvrb}
\usepackage{tabularx}  
\usepackage{threeparttable}
\usepackage{makecell}
\usepackage{booktabs}
\usepackage{adjustbox}
\usepackage{tcolorbox}
\usepackage{cleveref}
\usepackage{amsmath}

\usepackage{listings}
\usepackage{xcolor}

\title{Heterogeneous Dependency Graph-Guided Attention\\for Patent Representation Learning\\ }

\author{
Yongmin Yoo, \quad Qiongkai Xu, \quad Zhangkai Wu, \quad Longbing Cao \\
Frontier AI Research Centre, Macquarie University \\
School of Computing, FSE, Macquarie University\\
\texttt{yooyongmin91@gmail.com} \quad \\
\texttt{\{qiongkai.xu, zhangkai.wu, longbing.cao\}@mq.edu.au}
}

\begin{document}
\maketitle
\begin{abstract} 

Pre-trained language models advance patent classification and retrieval by encoding claims as flat token sequences, but they overlook the dependency hierarchy among claims. Incorporating this hierarchy into self-attention poses two challenges. First, claim dependencies include relation types with different levels of reliability, so treating them uniformly may allow noisy technical relations to interfere with more reliable legal citations. Second, claim dependencies are defined at the claim level, whereas Transformer attention operates over tokens, making direct structural injection nontrivial. We propose the Patent Heterogeneous Attention-Guided Graph Encoder (PHAGE), which constructs a typed claim graph that distinguishes legal citations from technical relations. PHAGE projects this claim-level topology into token-level attention through a connectivity mask and learnable relation-aware biases, and fine-tunes the encoder using a dual-granularity contrastive objective that combines inter-patent taxonomy with intra-patent topology. At inference, the graph-specific attention components are removed, allowing representations to be generated through a standard encoder forward pass without CDG construction. Experiments on patent classification, retrieval, and clustering show that PHAGE consistently outperforms domain-adapted and citation-aware baselines, demonstrating the value of claim-level structural guidance for graph-free patent representation learning.

\end{abstract}
\section{Introduction}

Patent claims form a directed dependency hierarchy in which dependent claims inherit limitations from antecedent claims and add narrower technical details~\citep{krestel2021survey}. Specifically, this structure diverges from the textual order: claims adjacent in a document are not necessarily related, while distant claims may be directly connected through dependency references.

PLM-based patent encoders, including BERT-for-Patents~\citep{srebrovic2020leveraging} and PatentBERT~\citep{lee2020patent}, have advanced patent analysis tasks such as classification and retrieval by capturing distributional semantics and domain-specific terminology. \footnote{A primer on patent claim structure is provided in Appendix~\ref{sec:appendix_primer} for readers unfamiliar with the domain.} However, these models operate through token-level contextual modeling, which captures semantic correlations among words but does not encode the explicit legal dependencies described above. Encoding claims as a flat sequence therefore conflates structurally proximate yet unrelated claims, blurring both legal scope and technical dependencies in the learned representation.


\begin{figure}[t]
    \centering
    \includegraphics[width=\linewidth]{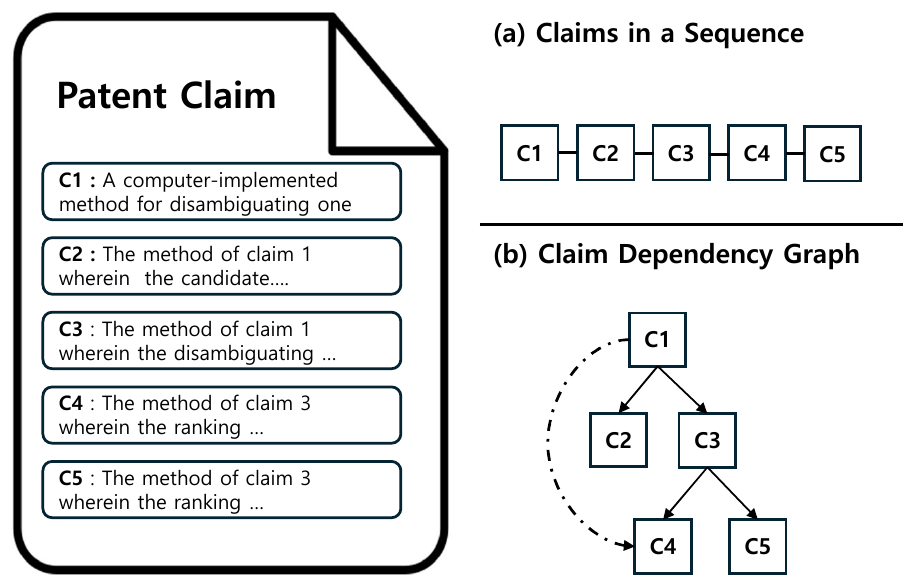}
    \caption{Comparison of patent claim encoding schemes. (a)~Sequential encoding treats claims as a flat token sequence, discarding inter-claim dependencies. (b)~PHAGE preserves the directed dependency hierarchy as a Claim Dependency Graph (CDG). Solid arrows denote direct citation edges; dashed arrows indicate transitive dependencies resolved via multi-hop attention.}
    \label{fig:intro_comparison}
\end{figure}



Recognizing this limitation, subsequent work has leveraged external signals such as IPC codes and citation networks as supervision~\citep{li2022copate, ghosh2024paecter} and explored graph-based architectures and section-level augmentation~\citep{bjorkqvist2023building, zuo2025patent}. However, these approaches use structure as training signal or auxiliary input at the document level, rather than injecting claim-level dependencies directly into the encoder's representation flow. We propose a patent representation model -- the \textbf{P}atent \textbf{H}eterogeneous \textbf{A}ttention-Guided \textbf{G}raph \textbf{E}ncoder (PHAGE),  which integrates claim-level dependency structure directly into self-attention.

PHAGE builds a heterogeneous claim graph through a deterministic pipeline that combines explicit claim references with rule-based technical relations, reducing reliance on learned edge prediction. During fine-tuning, it conditions self-attention on relation types through a connectivity mask and learnable relation-aware biases. The full encoder is optimized using a dual-granularity contrastive objective that combines patent-level supervision with claim-level structure. After training, the graph-specific attention components are removed, enabling the fine-tuned encoder to generate representations without CDG construction. Experiments on CLEF-IP, USPTO-70k, and DAPFAM show that PHAGE consistently outperforms strong domain-adapted and citation-aware baselines on patent classification, retrieval, and clustering under a frozen-encoder evaluation protocol.

PHAGE makes new contributions as follows:
\begin{itemize}
    \item \textbf{Intra-document claim topology as inductive bias.}~We identify that the directed dependency hierarchy among patent claims constitutes an informative signal that existing encoders discard, and then inject this topology directly into the encoder's representation flow rather than treating it as auxiliary input.

    \item \textbf{Topology-injected patent encoder.}~A novel model  PHAGE integrates claim-level dependencies into self-attention through a deterministic heterogeneous graph construction pipeline, a connectivity mask with learnable relation-aware biases, and a dual-granularity contrastive objective leveraging both patent-level taxonomy and claim-level topology.

    \item \textbf{Graph-free deployment after structure-guided training.}~PHAGE uses claim-level topology as training-time structural guidance while retaining the standard inference interface of the underlying patent encoder. Under graph-free downstream evaluation, PHAGE consistently outperforms the evaluated domain-adapted and citation-aware encoders across classification, retrieval, and clustering, demonstrating the practical value of intra-patent structural supervision without requiring claim graph construction at deployment.

\end{itemize}

\section{Related Work}

\subsection{Sequential Patent Encoders}
To address the complex syntax and vocabulary of technical literature, various Pre-trained Language Models tailored for the patent domain have been introduced. A foundational approach involves pre-training from scratch; for instance, BERT-for-Patents~\citep{srebrovic2020leveraging} leveraged over 100 million patent documents to capture the distributional semantics of technical terms, establishing a strong baseline.

Other strategies focus on domain adaptation via fine-tuning. PatentBERT~\citep{lee2020patent} optimized BERT-base specifically on claims for classification, while SciBERT~\citep{beltagy2019scibert} demonstrated effective transferability from scientific text, often enhanced by linguistically informed masking~\citep{althammer2021linguistically}. More recently, efficiency-oriented models have emerged: PatentSBERTa~\citep{bekamiri2024patentsberta} improved sentence representations for similarity search using silver labels, and ModernBERT has been adapted to process long-context technical disclosures more efficiently~\citep{yousefiramandi2025patent}.


Despite these advancements, conventional PLMs treat patent documents as linear text sequences, leaving the directed dependency structure among claims unmodeled.

\subsection{Inter-Document Graph Supervision}
To incorporate structural information beyond textual statistics, recent research has increasingly explored graph-based methodologies in patent analysis. A dominant stream focuses on modeling the macro-level citation network to refine document embeddings. For instance, PaECTER~\citep{ghosh2024paecter} and SearchFormer~\citep{vowinckel2023searchformer} utilize citation links as supervision signals, training models to pull cited patents closer in the vector space. More recently, PAI-NET~\citep{lee2025pai} extended this by incorporating prior art relationships and technical field hierarchies into a retrieval-augmented generation framework, further enhancing semantic search capabilities.


Beyond citation networks, graph-based indices linking claims to descriptions have been proposed for more structured retrieval~\citep{bjorkqvist2023building}. However, these methodologies primarily treat the patent document as a single node within a larger network or focus on external connectivity, largely overlooking the \textit{intra-document} claim dependency structure.

\subsection{Intra-Document Structure in Modeling}

Recent innovations seek to exploit fine-grained \textit{intra-document} features to advance patent representation learning. One direction involves data augmentation strategies that inject structural priors. \citet{zuo2025patent} introduce section-level augmentation to mitigate the embedding over-dispersion inherent in contrastive learning. Similarly, PatentMind~\citep{yoo2025patentmind} characterizes patents as multi-aspect reasoning graphs to enhance similarity evaluation.

Another line of work models claim dependencies using specialized graph networks. FLAN Graph~\citep{gao2024beyond} captures fine-grained claim dependencies to improve patent approval prediction, while HGM-Net~\citep{song2025research} integrates hierarchical contrastive learning with multimodal graph attention for feature fusion. However, these approaches largely operate through input-level augmentation, separate graph modules, or post-hoc modeling. None injects claim-level topology directly into the encoder's attention computation as an inductive bias. PHAGE addresses this gap by using the claim dependency hierarchy to guide self-attention during encoder fine-tuning.

\section{Methodology}

We propose the Patent Heterogeneous Attention-Guided Graph Encoder (PHAGE), which uses explicit claim structure during fine-tuning and performs graph-free encoding at inference; key notations used throughout this section are summarized in Appendix~\ref{app:notation}. First, we construct a heterogeneous Claim Dependency Graph (CDG) via a deterministic multi-signal extraction pipeline that represents explicit legal citations, technical term flows, and functional couplings as distinct edge types. Second, during fine-tuning, we incorporate the CDG into the self-attention mechanism of a pre-trained encoder through a connectivity mask and independently learnable biases conditioned on each edge type, thereby modulating information flow according to dependency semantics. Third, we optimize the full encoder using a dual-granularity contrastive objective that combines inter-document technological similarity with intra-document claim topology. The claim-level alignment signals are weighted by learnable relation-specific parameters, allowing the model to calibrate the contribution of each relation type during training. After fine-tuning, the CDG-derived mask, relation map, and relation-aware biases are omitted, and the resulting encoder processes each patent using a standard self-attention forward pass without graph construction.

\begin{figure*}[t]
\centering
\includegraphics[width=0.99\textwidth,keepaspectratio]{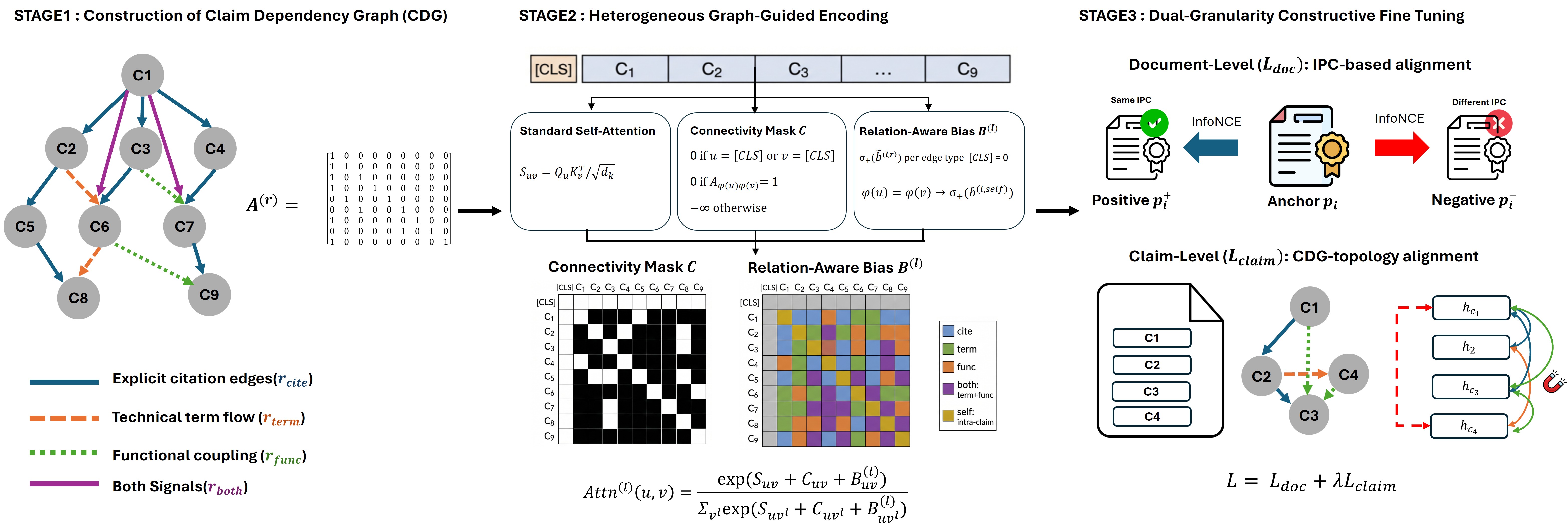}
\caption{Overview of the PHAGE Framework Pipeline. The framework consists of three stages: (1)~constructing a Claim Dependency Graph (CDG), (2)~encoding claims via a relation-guided structure-aware attention mechanism, and (3)~refining representations using a contrastive fine-tuning objective.}
\label{fig:phage_architecture}
\vspace{-4mm}
\end{figure*} 

\subsection{Construction of Heterogeneous Claim Dependency Graph}
\label{sec:cdg}
Patent claims form a directed dependency structure in which dependent claims inherit limitations from antecedent claims and may introduce additional technical restrictions. These dependencies are heterogeneous because explicit claim citations, term-level references, and functional couplings convey different semantics and may have different levels of extraction reliability. We represent this structure as a heterogeneous Claim Dependency Graph (CDG), denoted by $G=(V,E,\psi)$, where $V=\{c_1,\ldots,c_n\}$ is the set of claims in a patent, $E$ contains directed dependencies between distinct claims, and $\psi:E\rightarrow\mathbf{R}$ assigns each edge to one of four relation types:
\begin{equation}
\mathbf{R}
=
\left\{
r_{\mathrm{cite}},
r_{\mathrm{term}},
r_{\mathrm{func}},
r_{\mathrm{both}}
\right\}.
\label{eq:relation_set}
\end{equation}
All extracted edges are directed from an antecedent or defining claim toward a later dependent or referencing claim. This ordering preserves an acyclic claim-level structure. For an edge $(c_j,c_i,r)\in E$, claim $c_j$ provides antecedent information to claim $c_i$, and the corresponding attention convention allows query tokens in $c_i$ to attend to key tokens in $c_j$. Self-loops are excluded from $E$ and are introduced separately when constructing the attention mask.
\paragraph{Explicit Citation Extraction ($r_{\mathrm{cite}}$).} We extract the citation edge set $E_{\mathrm{cite}}$ from formal dependency declarations in claim preambles. Under common patent drafting conventions, a dependent claim explicitly identifies one or more antecedent claims through standardized expressions such as \textit{``The device of claim~1, wherein\ldots''}. We compile regular expressions matching the general schema $\langle\textit{prep}\rangle~\texttt{claim}~N$, where $\langle\textit{prep}\rangle$ ranges over \{\textit{of}, \textit{according to}, \textit{as in}, \textit{as claimed in}, \textit{as set forth in}\}, and $N$ denotes the referenced claim number. Each detected reference from claim $c_i$ to an antecedent claim $c_j$ produces a directed edge $(c_j,c_i,r_{\mathrm{cite}})$. We retain only directly stated citation edges and do not add their transitive closure to the graph. Indirect dependencies, such as claim~4 depending on claim~1 through claim~2, can instead be propagated across multiple encoder layers. To avoid assigning an implicit relation to a claim pair already connected by the explicit legal hierarchy, candidate implicit edges are excluded when their claim pair belongs to the transitive closure $E_{\mathrm{cite}}^{*}$.
\paragraph{Technical Term Flow Extraction ($r_{\mathrm{term}}$).} We construct $E_{\mathrm{term}}$ by tracking the introduction and subsequent reference of technical noun phrases across claims. Following common patent drafting patterns, a technical term is treated as introduced when it first occurs with an indefinite article such as \textit{a} or \textit{an}. A later occurrence with a definite article or demonstrative determiner, such as \textit{the} or \textit{said}, is treated as a reference to the earlier definition. When a term introduced in claim $c_j$ is subsequently referenced in claim $c_i$, we generate a candidate edge from $c_j$ to $c_i$. Because this procedure relies primarily on reference patterns rather than a fixed technical vocabulary, it can be applied to previously unseen technical terms. Candidate term-flow edges that duplicate a dependency represented in $E_{\mathrm{cite}}^{*}$ are excluded.
\paragraph{Functional Coupling Extraction ($r_{\mathrm{func}}$).} We construct $E_{\mathrm{func}}$ using domain-specific dependency patterns designed to capture functional relationships that may not be identified through article-based term tracking alone. We extract subject, verb, and object structures using a curated lexicon of 13 functional verbs, including \textit{couple}, \textit{connect}, and \textit{attach}. A candidate edge from claim $c_j$ to claim $c_i$ is created when a component involved in a functional construction in $c_i$ matches a technical term previously defined in $c_j$. Candidate functional edges that duplicate a dependency represented in $E_{\mathrm{cite}}^{*}$ are excluded. Additional extraction details are provided in Appendix~\ref{app:extraction}.
\paragraph{Heterogeneous Edge Integration.} We integrate the extracted relations while assigning each ordered claim pair to at most one edge type. Explicit citation edges are retained as $r_{\mathrm{cite}}$. For claim pairs without an explicit citation dependency, $r_{\mathrm{both}}$ is assigned when both term-flow and functional-coupling signals are present. A pair supported only by term flow is assigned $r_{\mathrm{term}}$, and a pair supported only by functional coupling is assigned $r_{\mathrm{func}}$. Consequently, the four relation types are mutually exclusive for every ordered pair of distinct claims. The resulting edge set contains only inter-claim dependencies:
\begin{equation}
E
\subseteq
\left\{
(c_j,c_i,r):
j\neq i,\ r\in\mathbf{R}
\right\}.
\label{eq:interclaim_edges}
\end{equation}
For each relation type $r\in\mathbf{R}$, we construct a typed adjacency matrix $\mathbf{A}^{(r)}\in\{0,1\}^{n\times n}$:
\begin{equation}
A^{(r)}_{ij}
=
\begin{cases}
1, & (c_j,c_i,r)\in E,\\
0, & \text{otherwise}.
\end{cases}
\label{eq:adjacency}
\end{equation}
Under this convention, $A^{(r)}_{ij}=1$ indicates that query tokens belonging to claim $c_i$ may receive information from key tokens belonging to claim $c_j$ through relation type $r$. The diagonal entries of all typed adjacency matrices are zero because intra-claim connectivity is represented separately rather than being assigned to each inter-claim relation type. This separation prevents self-loops from entering the claim-level contrastive objective and avoids assigning multiple relation labels to the same diagonal entry. Detailed extraction procedures and CDG statistics are provided in Appendices~\ref{app:extraction} and~\ref{app:cdg_stats}, respectively.

\subsection{Heterogeneous Graph-Guided Encoding}
\label{sec:encoding}
Standard Transformers apply unconstrained self-attention over token pairs without explicitly distinguishing the dependency relationships among patent claims. PHAGE instead uses the CDG as training-time structural guidance. It restricts direct cross-claim attention to extracted dependency edges and modulates permitted interactions according to their relation types. The resulting attention mechanism guides encoder fine-tuning, while all CDG-specific inputs and parameters are omitted during downstream inference.
\paragraph{Token-Level Projection.} The CDG is defined over claims, whereas Transformer attention operates over tokens. Let $\phi(u)$ denote the claim containing token $u$. Using the typed adjacency matrices $\{\mathbf{A}^{(r)}\}_{r\in\mathbf{R}}$ from \Cref{sec:cdg}, we construct a union adjacency matrix that combines inter-claim dependencies with intra-claim self-connectivity:
\begin{equation}
\bar{\mathbf{A}}
=
\mathbf{I}_n
\lor
\bigvee_{r\in\mathbf{R}}
\mathbf{A}^{(r)},
\label{eq:union_adjacency}
\end{equation}
where $\mathbf{I}_n$ is the $n\times n$ identity matrix and $\lor$ denotes element-wise logical disjunction. Thus, $\bar{A}_{ij}=1$ when $i=j$ or when at least one extracted dependency allows claim $c_i$ to receive information from claim $c_j$. During training, we project this claim-level structure into token-level attention using a connectivity mask $\mathbf{C}\in\mathbb{R}^{L\times L}$ and a relation-aware bias $\mathbf{B}^{(l)}\in\mathbb{R}^{L\times L}$ at encoder layer $l$, where $L$ is the input sequence length.
\paragraph{CDG-Guided Attention.} Let $\mathbf{S}^{\mathrm{attn}}_{uv}$ denote the scaled dot-product attention logit between query token $u$ and key token $v$:
\begin{equation}
\mathbf{S}^{\mathrm{attn}}_{uv}
=
\frac{\mathbf{Q}_u\mathbf{K}_v^\top}
{\sqrt{d_k}}.
\label{eq:attention_logit}
\end{equation}
During CDG-guided fine-tuning, the structural mask and relation-aware bias are added to this logit:
\begin{equation}
e_{uv}^{(l)}
=
\mathbf{S}^{\mathrm{attn}}_{uv}
+\mathbf{C}_{uv}
+\mathbf{B}^{(l)}_{uv}.
\label{eq:guided_logit}
\end{equation}
The resulting attention weight is:
\begin{equation}
\alpha_{uv,\mathrm{train}}^{(l)}
=
\frac{\exp\!\left(e_{uv}^{(l)}\right)}
{\displaystyle\sum_{v'=1}^{L}
\exp\!\left(e_{uv'}^{(l)}\right)}.
\label{eq:relation_attention}
\end{equation}
Entries assigned $-\infty$ in $\mathbf{C}$ receive zero attention probability because $\exp(-\infty)=0$, and the distribution is renormalized over the permitted key positions. The structural mask and relation-aware bias are applied only during CDG-guided fine-tuning.
\paragraph{Connectivity Mask.} The connectivity mask acts as a hard structural gate over direct cross-claim attention. Let $u_{\mathrm{cls}}$ denote the \texttt{[CLS]} token. The \texttt{[CLS]} query is permitted to attend to the complete input so that the document-level representation can aggregate information from all retained claims. Ordinary claim tokens may attend to tokens within the same claim or to tokens in an antecedent claim connected through the CDG. To prevent ordinary tokens from using \texttt{[CLS]} as an unrestricted cross-claim communication channel, attention from a non-\texttt{[CLS]} query to the \texttt{[CLS]} key is masked. The connectivity mask is defined as:
\begin{equation}
\mathbf{C}_{uv}
=
\begin{cases}
0, & u=u_{\mathrm{cls}},\\
-\infty, & v=u_{\mathrm{cls}},\\
0, & \bar{A}_{\phi(u)\phi(v)}=1,\\
-\infty, & \text{otherwise}.
\end{cases}
\label{eq:connectivity_mask}
\end{equation}
The cases are evaluated in the displayed order, so the first case applies when both $u$ and $v$ correspond to \texttt{[CLS]}. Because $\mathbf{I}_n$ is included in Eq.~\ref{eq:union_adjacency}, tokens belonging to the same claim remain mutually reachable. Cross-claim attention is permitted only when the corresponding ordered claim pair is connected in the CDG. The mean CDG density of 0.19 reported in Table~\ref{tab:cdg_basic} indicates that the mask removes a substantial proportion of direct cross-claim interactions while preserving intra-claim attention, CDG-supported cross-claim attention, and global aggregation through the \texttt{[CLS]} query.
\paragraph{Relation-Aware Bias.} The connectivity mask determines whether a token pair is reachable, whereas the relation-aware bias adjusts the relative attention strength assigned to permitted interactions. For each layer $l$, we introduce a raw scalar parameter $\tilde b^{(l,r)}$ for every relation type $r\in\mathbf{R}$ and a separate raw parameter $\tilde b^{(l,\mathrm{self})}$ for intra-claim interactions. Their positive transformed values are defined as:
\begin{equation}
\begin{aligned}
b^{(l,r)}
&=\sigma_+\!\left(\tilde b^{(l,r)}\right),\\
b^{(l,\mathrm{self})}
&=\sigma_+\!\left(\tilde b^{(l,\mathrm{self})}\right),
\end{aligned}
\label{eq:bias_transform}
\end{equation}
where $\sigma_+$ is the softplus function. This parameterization introduces five raw scalar parameters per layer, totaling 120 additional attention parameters for the 24-layer encoder. Because the edge integration procedure assigns each ordered pair of distinct claims to at most one relation type, the typed adjacency matrices satisfy:
\begin{equation}
\sum_{r\in\mathbf{R}}
A^{(r)}_{ij}
\leq 1,
\qquad i\neq j.
\label{eq:mutex}
\end{equation}
For a connected pair, let $\psi_{uv}=\psi(\phi(u),\phi(v))$ denote its relation type. For token pairs not involving \texttt{[CLS]}, the attention bias is:
\begin{equation}
\mathbf{B}^{(l)}_{uv}
=
\begin{cases}
b^{(l,\mathrm{self})},
& \phi(u)=\phi(v),\\
b^{(l,\psi_{uv})},
& A^{(\psi_{uv})}_{\phi(u)\phi(v)}=1,\\
0, & \text{otherwise}.
\end{cases}
\label{eq:relation_bias}
\end{equation}
For token pairs involving \texttt{[CLS]}, $\mathbf{B}^{(l)}_{uv}$ is set to zero. The final case in Eq.~\ref{eq:relation_bias} corresponds to a structurally disconnected pair and is masked by $\mathbf{C}$ during training. All raw parameters are initialized to zero, yielding $b^{(l,r)}=b^{(l,\mathrm{self})}=\ln 2\approx0.693$. Consequently, all permitted relation types begin with the same effective bias value, avoiding an initial preference for any particular dependency type.
\paragraph{Graph-Free Inference.} The connectivity mask, token-to-claim map, and relation-aware bias are used only during CDG-guided fine-tuning. Gradients from the document-level and claim-level objectives update the full encoder jointly with the additional relation-specific parameters. At inference, the CDG-specific components are omitted from the forward pass. This is mathematically equivalent to setting the structural mask and bias terms to zero, producing the standard self-attention distribution:
\begin{equation}
\alpha_{uv,\mathrm{infer}}^{(l)}
=
\frac{\exp\!\left(
\mathbf{S}^{\mathrm{attn}}_{uv}
\right)}
{\displaystyle\sum_{v'=1}^{L}
\exp\!\left(
\mathbf{S}^{\mathrm{attn}}_{uv'}
\right)}.
\label{eq:inference_attention}
\end{equation}
Inference therefore retains the fine-tuned encoder parameters but follows the standard self-attention computation of the underlying BERT architecture. The additional relation-specific parameters are not required after fine-tuning. This separation allows the CDG to provide structural supervision during representation learning without introducing graph construction or graph-specific computation at deployment.
\subsection{Dual-Granularity Contrastive Fine-tuning}
\label{sec:contrastive}
Although the backbone encoder provides domain-specific representations through masked language model pre-training, this objective does not explicitly capture inter-patent technological proximity or intra-patent claim dependencies. We therefore fine-tune the encoder using a dual-granularity contrastive objective that combines document-level IPC supervision with claim-level structural supervision derived from the CDG.
\paragraph{Document-Level Objective.} We use the International Patent Classification taxonomy to organize the document-level embedding space. Let $\{p_1,p_2,\ldots,p_N\}$ denote a batch of $N$ anchor patents. For each anchor patent $p_i$, we sample a positive patent $p_i^+$ from the same IPC subclass and a negative patent $p_i^-$ from a different IPC subclass. Positives and negatives are sampled uniformly from the training corpus without hard-negative mining. Let $\mathbf{H}_{p_i}$ denote the \texttt{[CLS]} representation of patent $p_i$. We define the positive and negative similarities as:
\begin{equation}
\begin{aligned}
s_i^+&=\cos\!\left(\mathbf{H}_{p_i},
\mathbf{H}_{p_i^+}\right),\\
s_i^-&=\cos\!\left(\mathbf{H}_{p_i},
\mathbf{H}_{p_i^-}\right).
\end{aligned}
\label{eq:doc_similarity}
\end{equation}
The document-level contrastive loss with temperature $\tau$ is:
\begin{equation}
\mathcal{L}_{\mathrm{doc}}
=-\frac{1}{N}\sum_{i=1}^{N}\ell_i^{\mathrm{doc}},
\label{eq:doc_loss}
\end{equation}
where
\begin{equation}
\ell_i^{\mathrm{doc}}
=
\log
\frac{\exp(s_i^+/\tau)}
{\exp(s_i^+/\tau)+\exp(s_i^-/\tau)}.
\label{eq:doc_instance_loss}
\end{equation}
\paragraph{Relation-Type-Aware Claim Alignment.} We introduce a claim-level objective grounded in the CDG to encourage structurally connected claims to have similar representations. Because the extracted relation types may differ in reliability and informativeness, we assign a learnable positive weight $w_r=\sigma_+(\tilde{w}_r)$ to each relation $r\in\mathbf{R}$, where $\sigma_+$ denotes the softplus function. All raw weights $\tilde{w}_r$ are initialized to zero, giving $w_r=\ln 2\approx0.693$ and an equal initial contribution across relation types. For a patent containing $n$ claims, let $\mathbf{h}_{c_i}$ denote the mean-pooled representation of claim $c_i$. We define the temperature-scaled similarity between claims $c_i$ and $c_j$ as:
\begin{equation}
S_{ij}
=
\frac{\cos\!\left(\mathbf{h}_{c_i},
\mathbf{h}_{c_j}\right)}{\tau_c},
\label{eq:claim_similarity}
\end{equation}
where $\tau_c$ is the claim-level temperature. For each directed dependency $(c_j,c_i,r)\in E$, the corresponding contrastive term is:
\begin{equation}
\ell_{ji}
=
-\log
\frac{\exp(S_{ji})}
{\displaystyle\sum_{\substack{k=1\\k\neq j}}^{n}
\exp(S_{jk})}.
\label{eq:claim_edge_loss}
\end{equation}
The relation-weighted claim-level objective is then:
\begin{equation}
\mathcal{L}_{\mathrm{claim}}
=
\frac{1}{|E|}
\sum_{(c_j,c_i,r)\in E}
w_r\,\ell_{ji}.
\label{eq:claim_loss}
\end{equation}
The edge set $E$ contains only inter-claim dependencies and excludes self-loops. For patents with no valid dependency edges after input truncation, we define $\mathcal{L}_{\mathrm{claim}}=0$, so these instances contribute only to the document-level objective. This formulation applies relation-specific alignment pressure to connected claims while allowing the model to calibrate the contribution of each extracted dependency type during training.
\paragraph{Joint Objective.} The final training objective combines the two levels of supervision:
\begin{equation}
\mathcal{L}
=
\mathcal{L}_{\mathrm{doc}}
+\lambda\mathcal{L}_{\mathrm{claim}},
\label{eq:joint_loss}
\end{equation}
where $\lambda$ controls the contribution of claim-level structural alignment. We select $\lambda=1.0$ based on validation over $\lambda\in\{0.1,0.5,1.0,2.0,5.0\}$, with stable validation performance observed for values between 0.5 and 2.0. Both contrastive objectives use a temperature of 0.05. During training, the full BERT-for-Patents backbone is optimized jointly with 124 additional scalar parameters, comprising 120 layer-specific attention biases and 4 relation-specific claim-loss weights. These additional parameters condition encoder fine-tuning on the CDG and are not used during graph-free inference. Downstream encoding retains the fine-tuned backbone and follows the standard self-attention computation in Eq.~\ref{eq:inference_attention}. Training details are provided in Appendix~\ref{app:training_details}, and computational complexity is analyzed in Appendix~\ref{app:complexity}. 
\section{Experiments}

We evaluate PHAGE on patent classification, retrieval, and clustering. PHAGE is initialized from BERT-for-Patents and fine-tuned on 131,653 USPTO patents filed from 2013 to 2017 using the dual-granularity objective and CDG-guided attention described in Section~\ref{sec:encoding}. During this training stage, the full backbone, the 120 relation-aware attention bias parameters, and the 4 relation-specific claim-loss weights are jointly optimized for 5 epochs using AdamW with a learning rate of $2\times10^{-5}$ on NVIDIA H100 GPUs. After training, the encoder parameters are frozen for all downstream evaluations. Neither the CDG connectivity mask nor the relation-aware attention bias is applied during downstream encoding, so PHAGE generates representations through a standard BERT forward pass without graph construction or auxiliary structural inputs. We verified through application-number deduplication that no patent in the training corpus appears directly in the evaluation sets. All comparison models are evaluated under the same frozen-encoder protocol, with task-specific linear probing for classification and fixed embedding-based evaluation for retrieval and clustering. Dataset statistics, full hyperparameters, and evaluation metrics are provided in Appendices~\ref{app:training_details}, \ref{app:training_data}, and~\ref{app:metrics}.

\subsection{Patent Classification}
\begin{table}[h]
\centering
\setlength{\tabcolsep}{4pt}
\resizebox{\columnwidth}{!}{%
\begin{tabular}{lcccc}
\toprule
\multirow{2}{*}{\textbf{Model}}
  & \multicolumn{2}{c}{\textbf{CLEF-IP}}
  & \multicolumn{2}{c}{\textbf{USPTO-70k}} \\
\cmidrule(lr){2-3}\cmidrule(lr){4-5}
 & Mi-F1$\uparrow$ & Ma-F1$\uparrow$ & Mi-F1$\uparrow$ & Ma-F1$\uparrow$ \\
\midrule
\multicolumn{5}{l}{\textbf{\textit{Sequence-only}}} \\
BERT-base          & 0.555 & 0.427 & 0.386 & 0.163 \\
SciBERT            & 0.417    & 0.260    & 0.347 & 0.133 \\
BERT-for-Patents   & 0.649 & 0.538 & 0.491 & 0.288 \\
\midrule
\multicolumn{5}{l}{\textbf{\textit{Citation-aware}}} \\
PatentSBERTa       & 0.586    & 0.476    & 0.456 & 0.253 \\
PaECTER            & 0.629 & 0.527 & 0.511 & 0.333 \\
PatEmbed-Base      & 0.630 & 0.549 & 0.523 & 0.317 \\
PatEmbed-Large     & 0.547 & 0.444 & 0.415 & 0.210 \\
\midrule
\multicolumn{5}{l}{\textbf{\textit{Graph-based}}} \\
GCN + BERT-for-Pat.    & 0.567    & 0.345    & 0.312    & 0.205    \\
GAT + BERT-for-Pat.    & 0.581    & 0.368    & 0.327    &  0.213    \\
GraphSAGE + BERT-for-Pat. &0.565 & 0.336   & 0.332    & 0.230    \\
\midrule
\multicolumn{5}{l}{\textbf{\textit{Claim-level graph}}} \\
\textbf{PHAGE (ours)} & \textbf{0.710} & \textbf{0.635} & \textbf{0.531} & \textbf{0.348} \\
\bottomrule
\end{tabular}}
\caption{Patent classification (linear probing, frozen encoder, 5-run avg.). Graph-based rows: BERT-for-Patents \texttt{[CLS]} with validation-tuned $k$-NN.}
\label{tab:classification}
\end{table}

We train a linear classifier on frozen \texttt{[CLS]} embeddings and report Micro-F1 and Macro-F1 averaged over five runs.~\Cref{tab:classification} reports classification results on CLEF-IP~\citep{piroi-etal-2011-clefip} and USPTO-70k~\citep{pujari-etal-2022-uspto70k}. PHAGE outperforms all evaluated baselines on both datasets, with the largest improvement observed in Macro-F1 on CLEF-IP, where it exceeds BERT-for-Patents by 9.7 points. Because Macro-F1 assigns equal importance to each class, this improvement is consistent with better representations of less frequent IPC subclasses, although class-level results are required to verify this interpretation directly. All three evaluated GNN variants degrade the performance of their respective backbones, suggesting that post-hoc message passing over task-agnostic embedding-similarity graphs may introduce neighborhood information that is not aligned with IPC class boundaries. However, because these GNN baselines operate on embedding-similarity graphs rather than the CDG used by PHAGE, the current comparison does not isolate the effects of graph construction and aggregation architecture. We therefore interpret these results as evidence that the evaluated similarity-graph pipelines are less effective than CDG-guided fine-tuning under our experimental setting, rather than as demonstrating a general limitation of graph neural networks.

\subsection{Patent Retrieval} 

\begin{table}[t]
\centering
\setlength{\tabcolsep}{5pt}
\begin{tabular}{lc}
\toprule
\textbf{Model} & \textbf{NDCG@100}$\uparrow$ \\
\midrule
\multicolumn{2}{l}{\textbf{\textit{Sequence-only}}} \\
BERT-base              & 0.170 \\
SciBERT                & 0.146 \\
BERT-for-Patents       & 0.228 \\
\midrule
\multicolumn{2}{l}{\textbf{\textit{Citation-aware}}} \\
PaECTER                & 0.343 \\
PatEmbed-Base          & 0.370 \\
PatEmbed-Large         & 0.377 \\
QaECTER\dag & 0.379 \\
\midrule
\multicolumn{2}{l}{\textbf{\textit{Graph-based}}} \\
GCN + BERT-for-Pat.       & 0.144 \\
GAT + BERT-for-Pat.       & 0.117 \\
GraphSAGE + BERT-for-Pat. & 0.084 \\
\midrule
\multicolumn{2}{l}{\textbf{\textit{Claim-level graph}}} \\
\textbf{PHAGE (ours)}  & \textbf{0.498} \\
\bottomrule
\end{tabular}
\caption{Patent retrieval on DAPFAM. Graph-based rows use BERT-for-Patents \texttt{[CLS]} with validation-tuned $k$-NN. \textsuperscript{\dag}Weights not public; score reported under the same DAPFAM evaluation protocol by~\citet{ayaou2025patenteb}.}
\label{tab:retrieval}
\end{table}

We encode queries and targets with the frozen encoder and rank candidates by cosine similarity, reporting NDCG@100.~\Cref{tab:retrieval} presents the results on DAPFAM, a cross-domain benchmark containing 1{,}247 query families and 45{,}336 target patents. PHAGE achieves an average NDCG@100 of 0.498 across six query--corpus configurations, as detailed in Appendix~\ref{app:dapfam_detail}. This result exceeds the best evaluated citation-aware baseline, QaECTER at 0.379, by 11.9 points and BERT-for-Patents at 0.228 by 27.0 points. The comparison is notable because PaECTER and PatEmbed are trained directly on examiner citation triplets designed for patent retrieval, whereas PHAGE uses no cross-patent citation links as supervision. PHAGE instead combines IPC-based inter-document supervision with structural guidance derived solely from intra-document claim dependencies. The evaluated GNN baselines also underperform their BERT-for-Patents backbone, suggesting that post-hoc message passing over embedding-similarity graphs may introduce neighborhoods that are not well aligned with retrieval relevance. Because PHAGE and the GNN baselines differ in both graph construction and integration architecture, these results should be interpreted as evidence that CDG-guided fine-tuning is more effective than the evaluated similarity-graph pipelines under our experimental protocol, rather than as establishing that graph semantics alone explain the observed performance gap.

\subsection{Patent Clustering}
\begin{table}[h]
\centering
\setlength{\tabcolsep}{4pt}
\resizebox{\columnwidth}{!}{%
\begin{tabular}{lcccc}
\toprule
\textbf{Model} & \textbf{NMI} & \textbf{ARI} & \textbf{Hom.} & \textbf{Com.} \\
\midrule
\multicolumn{5}{l}{\textbf{\textit{Sequence-only}}} \\
BERT-base        & 0.304 & 0.026 & 0.327 & 0.284 \\
SciBERT          & 0.303 & 0.029 & 0.325 & 0.284 \\
BERT-for-Pat.    & 0.431 & 0.058 & 0.466 & 0.401 \\
\midrule
\multicolumn{5}{l}{\textbf{\textit{Citation-aware}}} \\
PatentSBERTa     & 0.540    & 0.106    & 0.579    & 0.506    \\
PaECTER          & 0.565 & 0.099 & 0.614 & 0.524 \\
PatEmbed-Base    & 0.591 & 0.118 & 0.641 & 0.547 \\
PatEmbed-Large   & 0.229 & 0.013 & 0.242 & 0.217 \\
\midrule
\multicolumn{5}{l}{\textbf{\textit{Graph-based}}} \\
GCN + BERT-for-Pat.    &  0.351 & 0.035 & 0.379 & 0.327 \\
GAT + BERT-for-Pat.    & 0.365 & 0.048 & 0.393 & 0.341 \\
GraphSAGE + BERT-for-Pat. & 0.363 & 0.042 & 0.392 & 0.337 \\
\midrule
\multicolumn{5}{l}{\textbf{\textit{Claim-level graph}}} \\
\textbf{PHAGE (ours)} & \textbf{0.653} & \textbf{0.188} & \textbf{0.704} & \textbf{0.609} \\
\bottomrule
\end{tabular}}
\caption{Patent clustering on CLEF-IP (frozen encoder, $K{=}488$). Graph-based rows: BERT-for-Patents \texttt{[CLS]} with validation-tuned $k$-NN.}
\label{tab:clustering}
\end{table}

We apply K-Means to frozen \texttt{[CLS]} embeddings with $K{=}488$, corresponding to the number of IPC subclasses, and report NMI, ARI, Homogeneity, and Completeness.~\Cref{tab:clustering} presents clustering results on CLEF-IP without task-specific encoder fine-tuning. PHAGE achieves an NMI of 0.653, improving over BERT-for-Patents by 22.2 points. The improvement is larger for Homogeneity than for Completeness, at 23.8 and 20.8 points, respectively. This pattern suggests that PHAGE may improve intra-cluster purity more strongly than class-level coverage, although cluster-level error analysis would be required to verify this interpretation. As in classification, post-hoc GNN processing degrades the BERT-for-Patents representations under the evaluated embedding-similarity graph setting. This result suggests that the resulting $k$-NN neighborhoods may not align well with IPC subclass structure, but it does not isolate whether the degradation arises from graph construction, message-passing architecture, or their interaction.

\subsection{Ablation Study}
\label{sec:ablation}

\begin{table}[h]
\centering
\small
\setlength{\tabcolsep}{7pt}
\begin{tabular}{lccc}
\toprule
\multirow{2}{*}{\textbf{Variant}}
  & \textbf{Classif.} & \textbf{Retrieval} & \textbf{Cluster.} \\
  & \textbf{Ma-F1} & \textbf{NDCG@100} & \textbf{NMI} \\
\midrule
BERT-for-Pat.     & 0.538 & 0.228 & 0.431 \\
\midrule
\multicolumn{4}{l}{\textbf{\textit{CDG edge types}}} \\
No Graph                        & 0.582 & 0.371 & 0.523 \\
$r_{\text{cite}}$ only          & 0.611 & 0.432 & 0.587 \\
w/o $r_{\text{term}}$           & 0.623 & 0.479 & 0.636 \\
w/o $r_{\text{func}}$           & 0.626 & 0.484 & 0.641 \\
w/o $r_{\text{both}}$           & 0.619 & 0.473 & 0.631 \\
\midrule
\multicolumn{4}{l}{\textbf{\textit{Attention components}}} \\
w/o mask $\mathcal{C}$          & 0.607 & 0.441 & 0.594 \\
w/o bias $\mathcal{B}$          & 0.618 & 0.462 & 0.621 \\
w/o both                        & 0.586 & 0.378 & 0.530 \\
\midrule
\multicolumn{4}{l}{\textbf{\textit{Contrastive objective}}} \\
$\mathcal{L}_{\text{doc}}$ only   & 0.614 & 0.412 & 0.571 \\
$\mathcal{L}_{\text{claim}}$ only & 0.573 & 0.449 & 0.608 \\
Uniform $w_r$                     & 0.628 & 0.487 & 0.643 \\
\midrule
\textbf{PHAGE (full)}           & \textbf{0.635}
                                & \textbf{0.498}
                                & \textbf{0.653} \\
\bottomrule
\end{tabular}
\caption{Ablation study isolating CDG edge types, attention components, and contrastive objective variants (CLEF-IP Ma-F1 / DAPFAM NDCG@100 / CLEF-IP NMI). BERT-for-Patents: unmodified backbone.}
\label{tab:ablation}
\end{table}

\Cref{tab:ablation} isolates the contribution of each PHAGE component across three axes; all variants share the same backbone and training setup.

\paragraph{CDG edge types.} The improvement from BERT-for-Patents to \textit{No Graph} (0.538 $\to$ 0.582) reflects the effect of document-level contrastive fine-tuning without CDG-guided attention or claim-level structural alignment. The further improvement from \textit{No Graph} to PHAGE (0.582 $\to$ 0.635) measures the combined contribution of CDG-guided attention and the claim-level objective beyond document-level contrastive fine-tuning. The $r_{\mathrm{cite}}$-only variant increases Macro-F1 from 0.582 to 0.611, suggesting that explicit citation dependencies provide useful structural supervision even without the implicit relation types. The implicit edge ablations further reveal that edge frequency does not directly determine downstream contribution. Although $r_{\mathrm{func}}$ accounts for 60.3\% of all extracted edges, removing it causes a smaller performance reduction than removing $r_{\mathrm{term}}$, which accounts for 7.9\% of the edges. Similarly, $r_{\mathrm{both}}$ represents only 2.6\% of the extracted edges but produces a comparatively large ablation effect. These results suggest that dependencies supported by both term-flow and functional-coupling signals form a particularly informative subset, although direct annotation-based validation is required before interpreting this subset as having higher extraction precision.

\paragraph{Attention components.} Removing the connectivity mask reduces Macro-F1 from 0.635 to 0.607, whereas removing the relation-aware bias reduces it to 0.618. Similar patterns are observed in retrieval and clustering. These results suggest that the connectivity mask contributes more strongly than relation-specific bias differentiation under the evaluated ablation setting. The \textit{w/o both} variant removes both attention components but retains the CDG for claim-level contrastive supervision, whereas \textit{No Graph} removes both CDG-guided attention and the claim-level objective. The \textit{w/o both} variant performs close to \textit{No Graph} in Macro-F1 (0.586 vs.\ 0.582), suggesting that a substantial portion of the observed structural gain is associated with attention injection. The small numerical difference is consistent with an additional contribution from the claim-level objective, although variability estimates are required to determine whether this difference is statistically reliable.

\paragraph{Contrastive objective.} The two contrastive objectives exhibit different task-level effects. Using $\mathcal{L}_{\mathrm{doc}}$ alone yields a Macro-F1 of 0.614 but an NDCG@100 of 0.412, whereas using $\mathcal{L}_{\mathrm{claim}}$ alone yields a lower Macro-F1 of 0.573 but a higher NDCG@100 of 0.449. The claim-level objective also produces a higher clustering NMI than the document-level objective, at 0.608 versus 0.571. These contrasting patterns suggest that the objectives provide complementary supervision: the document-level objective contributes more strongly to IPC classification, whereas the claim-level objective performs better on retrieval and clustering when each objective is evaluated independently. The improvement from \textit{No Graph} at 0.582 to $\mathcal{L}_{\mathrm{doc}}$ only at 0.614 further indicates that CDG-guided attention contributes beyond document-level contrastive fine-tuning, since neither variant includes the claim-level objective. Learnable relation weights $w_r$ provide a small but consistent improvement over uniform weighting across the reported tasks, with their converged values analyzed in \Cref{sec:discussion}.

\section{Discussion}
\label{sec:discussion}

\paragraph{Finding 1: Intra-document claim topology provides complementary structural supervision.} PHAGE outperforms the evaluated citation-aware baselines across classification, retrieval, and clustering without using cross-patent citation links as structural supervision. The \textit{No Graph} variant, which retains document-level contrastive fine-tuning but removes CDG-guided attention and claim-level structural alignment, improves CLEF-IP Macro-F1 from 0.538 to 0.582. PHAGE further increases Macro-F1 to 0.635, indicating that CDG-guided training provides gains beyond those obtained from the shared document-level contrastive objective. Consistent improvements in retrieval and clustering further suggest that intra-document claim topology provides complementary structural supervision under our evaluation protocol. These comparisons do not establish that intra-document topology is inherently superior to cross-patent citation structure, but they demonstrate that PHAGE is competitive with the evaluated citation-aware encoders without requiring cross-patent citation graphs.

\paragraph{Finding 2: Learned relation biases exhibit depth-dependent patterns.} The learned relation-aware biases vary across both relation type and encoder depth. As reported in Appendix~\ref{app:learned_params}, layers 0--9 assign larger average biases to the implicit relations, with values of 0.6964 for $r_{\mathrm{term}}$ and 0.6954 for $r_{\mathrm{func}}$, than to the explicit citation relation at 0.6905. In contrast, layers 18--23 assign a larger average bias to the intra-claim relation, increasing from 0.6921 in layers 0--9 to 0.7030 in layers 18--23, while the average cross-claim relation biases decrease. These descriptive patterns suggest that lower layers place relatively greater emphasis on cross-claim relational information, whereas upper layers increasingly favor information exchange within individual claims. The ablation results provide complementary evidence that relation-aware attention contributes to downstream performance: removing the bias term reduces Macro-F1 from 0.635 to 0.618, retrieval NDCG@100 from 0.498 to 0.462, and clustering NMI from 0.653 to 0.621. Replacing the learned claim-loss weights with uniform weights produces smaller reductions, yielding 0.628 Macro-F1, 0.487 NDCG@100, and 0.643 NMI. Under the evaluated ablation setting, these results suggest that layer-specific attention modulation contributes more strongly than relation-specific weighting of the claim-level objective. Because the learned bias differences are numerically modest, we interpret their depth-dependent variation as a descriptive pattern rather than conclusive evidence of functional specialization.

\paragraph{Finding 3: Training-time structural guidance improves representations used for graph-free inference.} 
PHAGE applies the CDG connectivity mask and relation-aware bias during fine-tuning but omits both components during downstream encoding. Consequently, all reported PHAGE results are obtained from a standard forward pass of the fine-tuned backbone without CDG construction, token-to-claim relation mapping, or relation-specific attention inputs. Under this graph-free evaluation protocol, PHAGE achieves a Macro-F1 of 0.635, compared with 0.582 for the \textit{No Graph} variant. Because the two variants share the document-level contrastive objective but differ in both CDG-guided attention and claim-level structural alignment, this comparison measures the combined contribution of CDG-guided training rather than the isolated effect of either component. The corresponding improvements in retrieval and clustering indicate that this benefit is observed across multiple downstream tasks. The comparison with the \textit{w/o both} variant further suggests that training-time attention injection provides gains beyond claim-level contrastive supervision under the evaluated ablation setting. However, the current experiments do not compare graph-free and graph-conditioned inference using the same PHAGE checkpoint. The results therefore demonstrate the effectiveness of graph-free representations produced after CDG-guided fine-tuning, but they do not establish how much performance is retained relative to graph-conditioned inference or whether removing the CDG at inference incurs no performance loss.

\section{Conclusion}
We presented PHAGE, a patent representation model that uses a heterogeneous claim dependency graph to guide Transformer attention during fine-tuning. PHAGE combines a structural connectivity mask, layer-specific relation-aware biases, and a dual-granularity contrastive objective to incorporate both inter-patent taxonomic supervision and intra-patent claim topology. Across patent classification, retrieval, and clustering, the resulting encoder outperforms the evaluated domain-adapted and citation-aware baselines under a frozen-encoder evaluation protocol. Importantly, downstream representations are generated without constructing the CDG or applying graph-specific attention components, allowing PHAGE to retain the standard inference interface of its BERT backbone. The results indicate that claim-level dependency structure provides complementary supervision beyond document-level contrastive learning and can improve representations used for graph-free downstream encoding. Future work will evaluate graph-conditioned inference using the same trained checkpoint, validate the extracted dependency relations across languages and patent authorities, and investigate whether training-time structural guidance transfers to other hierarchically organized technical documents.

\section*{Limitations}
\label{sec:limitations}

This work focuses on claim dependency structures in English-language patents. The technical term-flow component $r_{\text{term}}$ relies on English article transitions, including the introduction of a term with an indefinite article and its subsequent reference with a definite article or demonstrative determiner. Languages without comparable article systems, including Chinese, Japanese, and Korean, require language-specific methods for identifying term introduction and reference. Although $r_{\text{term}}$ accounts for 7.9\% of the extracted CDG edges, the citation and functional-coupling rules may also require adaptation to differences in patent drafting conventions, syntactic structure, and available parsing resources. The attention mechanism itself operates on abstract claim-level adjacency matrices and does not assume a specific language, but the cross-lingual portability of the complete extraction pipeline remains to be empirically evaluated.

PHAGE uses the CDG connectivity mask and relation-aware bias during fine-tuning but removes both components during downstream inference. This creates a deliberate difference between the attention computation used for representation learning and the standard self-attention computation used for deployment. The design eliminates the need to construct a claim graph for each new patent and preserves compatibility with conventional BERT-based embedding pipelines. The reported results show that a backbone trained with CDG guidance outperforms the corresponding No Graph variant when both are evaluated without graph inputs. However, this comparison evaluates models trained under different structural conditions and does not directly measure the performance difference between graph-conditioned and graph-free inference for the same checkpoint. A controlled comparison that enables and disables CDG injection at inference is therefore required to quantify the trade-off between predictive performance and deployment efficiency. The present study also verifies overlap using application numbers, which may not identify all related applications, continuation filings, or members of the same international patent family. Future work should adopt family-level deduplication and temporal evaluation to provide a stricter assessment of generalization.

\section*{Ethical Considerations}
During the preparation of this work, the author(s) utilized generative AI to refine linguistic clarity and support the creation of certain diagrams. The author(s) carefully reviewed all outputs and maintain full responsibility for the intellectual content and originality of the final paper.

\clearpage
\newpage
\bibliography{custom}
\bibliographystyle{acl_natbib}

\appendix
\clearpage
\newpage
\appendix
\crefname{section}{Appendix}{Appendices}

\section{Patent Claim Structure Primer}
\label{sec:appendix_primer}

This appendix provides background on patent documents and claim structure for readers unfamiliar with the patent domain.

\paragraph{What is a Patent?}
A patent is a legal document granted by a government authority (e.g., the United States Patent and Trademark Office) that gives an inventor exclusive rights to an invention for a limited period, typically 20 years. A patent document consists of several sections: a title, abstract, detailed description, drawings, and claims. Among these, the claims section is the most legally significant, as it defines the precise scope of legal protection.

\paragraph{What is a Patent Claim?}
A patent claim is a single sentence that defines the legal boundary of an invention's protection. Every granted patent contains one or more claims, and it is the claims, not the description or drawings, that determine what is legally protected.

\paragraph{Independent vs.\ Dependent Claims.}
Claims are of two types. An \textit{independent claim} stands alone and defines the broadest scope of protection (e.g., ``A method comprising steps X, Y, and Z.''). A \textit{dependent claim} explicitly references a prior claim and adds narrower limitations (e.g., ``The method of claim 1, wherein step X uses algorithm A.''). Dependent claims inherit all limitations of the claim(s) they reference.

\paragraph{Dependency Hierarchy.}
These references form a directed acyclic graph (DAG). For example, given five claims:
\begin{itemize}
    \item C1: independent claim (broadest scope)
    \item C2: depends on C1
    \item C3: depends on C1
    \item C4: depends on C3
    \item C5: depends on C3
\end{itemize}
The resulting hierarchy is a tree rooted at C1, where C4 and C5 inherit limitations from both C3 and C1. Crucially, this structure is \textit{not} reflected in textual order: C2 and C3 are adjacent in the document but share no dependency, while C4 and C1 are distant but legally linked through C3.

\paragraph{Why This Matters for Encoding.}
Standard language models encode claims as a flat token sequence, treating all claims as equally related. This discards the dependency hierarchy, conflating structurally connected claims (e.g., C4 and C3) with unrelated ones (e.g., C2 and C4). PHAGE addresses this by injecting the claim dependency graph directly into the encoder's attention mechanism.

\section{Computational Complexity Analysis}
\label{app:complexity}

Standard self-attention requires $O(L^2d)$ operations per layer, where $L$ is the input sequence length and $d$ is the hidden dimension. During CDG-guided fine-tuning, PHAGE adds two element-wise operations to the pre-softmax attention logits: a connectivity mask $\mathbf{C}\in\mathbb{R}^{L\times L}$ derived from the CDG and a relation-aware bias $\mathbf{B}^{(l)}\in\mathbb{R}^{L\times L}$ constructed by broadcasting 5 layer-specific scalar parameters according to a precomputed token-to-claim relation map. Constructing and applying these matrices requires $O(L^2)$ operations per layer, which does not change the asymptotic $O(L^2d)$ complexity of self-attention. CDG extraction and token-to-claim mapping are performed as preprocessing steps for the training corpus and are not required during downstream inference.

The claim-level contrastive objective $\mathcal{L}_{\text{claim}}$ computes pairwise cosine similarities among $n$ claim representations, requiring $O(n^2d)$ operations per patent. Because the number of claims is substantially smaller than the token sequence length in our corpus, with a mean of 16.85 claims and a maximum input length of 512 tokens, this computation is smaller than the encoder's self-attention cost.

At inference, PHAGE removes both the CDG connectivity mask and the relation-aware bias and retains only the fine-tuned BERT-for-Patents backbone. Its inference-time asymptotic complexity is therefore identical to that of the underlying encoder, namely $O(L^2d)$ per layer, and no additional graph construction or graph-specific attention computation is required.

\section{Notation Summary}
\label{app:notation}

We summarize and identify all symbols that we employed in~\Cref{tab:notation}.

\begin{table}[h]
\centering
\small
\caption{Summary of key notations.}
\label{tab:notation}
\setlength{\tabcolsep}{3pt}
\resizebox{\columnwidth}{!}{%
\begin{tabular}{lll}
\toprule
\textbf{Symbol} & \textbf{Shape} & \textbf{Meaning} \\
\midrule
$G = (V, E, \psi)$ & & Claim Dependency Graph \\
$V = \{c_1, \ldots, c_n\}$ & & Claims in a patent \\
$\psi: E \to \mathbf{R}$ & & Edge type function \\
$\mathbf{R}$ & $\{r_\text{cite}, r_\text{term}, r_\text{func}, r_\text{both}\}$ & Relation types \\
$\mathbf{A}^{(r)}$ & $\{0,1\}^{n \times n}$ & Typed adjacency \\
$\bar{\mathbf{A}}$ & $\{0,1\}^{n \times n}$ & Union adjacency \\
$\phi(u)$ & $\to \{1,\ldots,n\}$ & Token-to-claim map \\
$\mathbf{C}$ & $\mathbb{R}^{L \times L}$ & Connectivity mask \\
$\mathbf{B}^{(l)}$ & $\mathbb{R}^{L \times L}$ & Layer-$l$ bias matrix \\
$b^{(l,r)}, b^{(l,\text{self})}$ & $\mathbb{R}$ & Scalar biases (5/layer) \\
$\mathbf{H}_{p_i}$ & $\mathbb{R}^{d}$ & \texttt{[CLS]} of patent $p_i$ \\
$\mathbf{h}_{c_i}$ & $\mathbb{R}^{d}$ & Mean-pooled claim $c_i$ \\
$w_r$ & $\mathbb{R}_{+}$ & Contrastive weight \\
$\tau, \tau_c$ & $\mathbb{R}_{+}$ & Temperatures \\
$\lambda$ & $\mathbb{R}_{+}$ & Claim loss weight \\
\bottomrule
\end{tabular}}
\end{table}

\section{Multi-Signal Extraction Details}
\label{app:extraction}

\paragraph{Technical Term Flow ($r_{\text{term}}$).}
We identify definition loci of technical terms via noun phrase extraction. A stop-term list $\mathcal{S}$ filters ubiquitous expressions (e.g., \textit{the method}, \textit{the system}), and terms in more than $\rho{=}0.8$ of claims are excluded. The definition locus $\textsc{def}(t)$ is the first claim where $t$ appears with an indefinite article. Edges connect this locus to claims referencing $t$ via definite articles or demonstrative determiners, excluding pairs in the transitive closure $E_{\text{cite}}^{*}$:

\begin{equation}
\begin{aligned}
E_{\text{term}} = \{ (c_j, c_i, r_{\text{term}}) \mid{} & \exists\, t \in T(c_i) : \textsc{def}(t) = c_j, \\
& \text{$t$ has def./dem.\ ref.\ in $c_i$}, \\
& c_j \neq c_i,\; (c_j, c_i) \notin E_{\text{cite}}^{*} \}
\end{aligned}
\end{equation}

\paragraph{Functional Coupling ($r_{\text{func}}$).}
A verb lexicon $\mathcal{R}_{\text{lex}}$ of 13 functional verbs (\textit{couple, connect, attach, mount, dispose, position, support, engage, receive, extend, interpose, configure, adapt}) captures predicate--argument relations. We extract subject--object pairs via dependency parsing, retaining at most $K{=}8$ components per claim. An edge is added when a component matches a term defined in another claim:

\begin{equation}
\begin{aligned}
E_{\text{func}} = \{ (c_j, c_i, r_{\text{func}}) \mid{} & \exists\, r \in \mathcal{R}_{\text{lex}},\, e \in \textsc{comp}(c_i) : \\
& e \in T(c_j),\; c_j \neq c_i, \\
& (c_j, c_i) \notin E_{\text{cite}}^{*} \}
\end{aligned}
\end{equation}

\section{CDG Statistics}
\label{app:cdg_stats}

All statistics are computed over 131,653 training patents.

\begin{table}[h]
\centering
\small
\caption{Basic CDG statistics. Path length and diameter from 10K sampled multi-claim patents.}
\label{tab:cdg_basic}
\resizebox{\columnwidth}{!}{%
\begin{tabular}{lcccc}
\toprule
& \textbf{Mean} & \textbf{Std} & \textbf{Med.} & \textbf{Range} \\
\midrule
Claims/patent     & 16.85 & 8.60  & 18   & [1, 116] \\
Edges/patent      & 52.63 & 71.00 & 32   & [0, 4{,}907] \\
Density           & 0.19  & 0.13  & 0.20 & [0, 1.0] \\
Avg.\ path length & 2.01  & 0.50  & 1.93 & --- \\
Diameter          & 3.68  & 1.41  & 3.00 & --- \\
\bottomrule
\end{tabular}}
\end{table}

\begin{table}[h]
\centering
\small
\caption{Edge distribution by relation type.}
\label{tab:cdg_edge_dist}
\resizebox{\columnwidth}{!}{%
\begin{tabular}{lrrrc}
\toprule
\textbf{Type} & \textbf{Count} & \textbf{\%} & \textbf{Mean/pat.} & \textbf{Presence} \\
\midrule
$r_{\text{cite}}$ & 2{,}025{,}334 & 29.2 & 15.38 & 97.5\% \\
$r_{\text{term}}$ &   546{,}225 &  7.9 &  4.15 & 65.8\% \\
$r_{\text{func}}$ & 4{,}179{,}022 & 60.3 & 31.74 & 72.2\% \\
$r_{\text{both}}$ &   178{,}325 &  2.6 &  1.35 & 37.7\% \\
\midrule
Total & 6{,}928{,}906 & 100.0 & 52.63 & --- \\
\bottomrule
\end{tabular}}
\end{table}

\paragraph{Boundary cases.}
Of 131,653 patents, 3,029 (2.3\%) have zero edges (2,438 are single-claim). The remaining 591 default to unconstrained attention. An additional 16,252 (12.3\%) have only $r_{\text{cite}}$ edges, so implicit extraction provides complementary signal for 85.3\% of the corpus.

\section{Learned Parameter Analysis}
\label{app:learned_params}

\paragraph{Contrastive weights.}
All converged weights are near-uniform ($w_r \approx 0.673$), indicating the model does not rely on differential loss weighting.

\begin{table}[h]
\centering
\small
\caption{Learned contrastive weights after convergence.}
\label{tab:relation_weights}
\begin{tabular}{lcc}
\toprule
\textbf{Relation} & $\tilde{w}_r$ & $w_r = \sigma^+(\tilde{w}_r)$ \\
\midrule
$r_{\text{cite}}$ & $-0.0415$ & $0.6726$ \\
$r_{\text{both}}$ & $-0.0413$ & $0.6727$ \\
$r_{\text{term}}$ & $-0.0411$ & $0.6728$ \\
$r_{\text{func}}$ & $-0.0409$ & $0.6729$ \\
\bottomrule
\end{tabular}
\end{table}

\paragraph{Layer-wise attention biases.}
Layers 0--9 assign larger average biases to the implicit relations ($r_{\mathrm{term}}$: 0.6964, $r_{\mathrm{func}}$: 0.6954) than to citations (0.6905). Layers 18--23 assign a larger average bias to the intra-claim relation than layers 0--9 (0.7030 vs.\ 0.6921), while assigning smaller average biases to the cross-claim relations. These values reveal a depth-dependent descriptive pattern, but they do not independently establish functional specialization across layers.

\begin{table}[h]
\centering
\small
\caption{Learned effective attention biases $b^{(l,r)}=\sigma_+\!\left(\tilde b^{(l,r)}\right)$ across encoder layers. The values exhibit descriptive variation across relation types and depth.}
\label{tab:bias_layerwise}
\setlength{\tabcolsep}{4pt}
\resizebox{\columnwidth}{!}{%
\begin{tabular}{rccccc}
\toprule
\textbf{Lay.} & \textbf{self} & $r_{\text{cite}}$ & $r_{\text{term}}$ & $r_{\text{func}}$ & $r_{\text{both}}$ \\
\midrule
0  & .6907 & .6919 & .6950 & .6956 & .6932 \\
1  & .6921 & .6915 & .6946 & .6948 & .6942 \\
2  & .6902 & .6890 & .6981 & .6983 & .6944 \\
3  & .6930 & .6894 & .6955 & .6952 & .6933 \\
4  & .6920 & .6903 & .6952 & .6964 & .6940 \\
5  & .6928 & .6908 & .6944 & .6951 & .6936 \\
6  & .6902 & .6897 & .6990 & .6983 & .6948 \\
7  & .6943 & .6877 & .7001 & .6931 & .6933 \\
8  & .6929 & .6912 & .6968 & .6944 & .6929 \\
9  & .6929 & .6931 & .6948 & .6927 & .6935 \\
\midrule
10 & .6930 & .6958 & .6925 & .6913 & .6921 \\
11 & .6959 & .6916 & .6914 & .6910 & .6923 \\
12 & .6914 & .6890 & .7013 & .6963 & .6934 \\
13 & .6929 & .6909 & .6957 & .6944 & .6935 \\
14 & .6896 & .6923 & .6967 & .6973 & .6948 \\
15 & .6940 & .6924 & .6970 & .6923 & .6928 \\
16 & .6954 & .6895 & .6930 & .6923 & .6922 \\
17 & .6946 & .6925 & .6900 & .6918 & .6939 \\
\midrule
18 & .6991 & .6866 & .6892 & .6898 & .6924 \\
19 & .7033 & .6869 & .6873 & .6836 & .6887 \\
20 & .7035 & .6874 & .6869 & .6823 & .6890 \\
21 & .7052 & .6870 & .6855 & .6792 & .6870 \\
22 & .7047 & .6877 & .6850 & .6799 & .6861 \\
23 & .7023 & .6894 & .6876 & .6842 & .6871 \\
\midrule
\multicolumn{6}{l}{\textit{Group means}} \\
0--9   & .6921 & .6905 & .6964 & .6954 & .6937 \\
18--23 & .7030 & .6875 & .6869 & .6832 & .6884 \\
\bottomrule
\end{tabular}}
\end{table}

\section{Evaluation Metrics}
\label{app:metrics}

\paragraph{Classification.} We report Micro-F1 and Macro-F1. Micro-F1 aggregates true positives, false positives, and false negatives across all classes, favoring majority classes. Macro-F1 averages per-class F1 scores equally, reducing the dominance of frequent classes in the aggregate evaluation.

\paragraph{Retrieval.} We report NDCG@100 (Normalized Discounted Cumulative Gain at rank 100), which measures ranking quality by assigning higher credit to relevant documents appearing at top positions.

\paragraph{Clustering.} We report four metrics: Normalized Mutual Information (NMI), which measures the agreement between predicted clusters and ground-truth labels normalized by entropy; Adjusted Rand Index (ARI), which counts pairwise agreement corrected for chance; Homogeneity, which measures whether each cluster contains only members of a single class; and Completeness, which measures whether all members of a class are assigned to the same cluster.

\section{Training Hyperparameters}
\label{app:training_details}

Table~\ref{tab:hyperparams} summarizes the training configuration of PHAGE. We initialize the encoder with BERT-for-Patents (Large) and fine-tune the full backbone jointly with 120 layer-specific relation-aware attention biases and 4 relation-specific claim-loss weights. The model is trained for 5 epochs using AdamW with a learning rate of $2\times10^{-5}$ and gradient accumulation over 128 steps, yielding an effective batch size of 512 triplets. The document-level and claim-level temperatures are both set to 0.05. The claim-loss coefficient $\lambda$ is set to 1.0 based on validation over $\lambda\in\{0.1,0.5,1.0,2.0,5.0\}$. CDG-derived connectivity masks and relation-aware biases are used during training only. During downstream evaluation, the fine-tuned encoder is frozen, and representations are extracted without constructing the CDG or applying the structural attention components. Training is conducted on NVIDIA H100 GPUs.

\begin{table}[h]
\centering
\small
\caption{Training and inference configuration for PHAGE.}
\label{tab:hyperparams}
\begin{tabular}{ll}
\toprule
\textbf{Parameter} & \textbf{Value} \\
\midrule
Backbone & BERT-for-Patents (Large) \\
Backbone optimization & Full fine-tuning \\
Hidden size & 1{,}024 \\
Layers / heads & 24 / 16 \\
Maximum sequence length & 512 \\
\midrule
Optimizer & AdamW \\
Learning rate & $2\times10^{-5}$ \\
Weight decay & 0.01 \\
Gradient clipping & Maximum norm 1.0 \\
Batch size & 4 triplets \\
Gradient accumulation & 128 steps \\
Effective batch size & 512 triplets \\
Epochs & 5 \\
\midrule
$\tau$ for document loss & 0.05 \\
$\tau_c$ for claim loss & 0.05 \\
$\lambda$ & 1.0 \\
Additional attention biases & 120, 5 per layer \\
Additional claim-loss weights & 4 \\
\midrule
CDG used during training & Yes \\
CDG used during inference & No \\
Downstream encoder update & Frozen \\
\bottomrule
\end{tabular}
\end{table}

\section{Training Corpus}
\label{app:training_data}

PHAGE is trained on 131,653 USPTO patents filed between 2013 and 2017, sourced from the Harvard USPTO Patent Dataset~\citep{suzgun2023hupd}. This dataset provides full-text patent applications including titles, abstracts, claims, and metadata such as International Patent Classification (IPC) codes. We use this publicly available corpus to ensure reproducibility.

\paragraph{Corpus Composition.}
The corpus spans 610 unique IPC subclasses and 36,094 unique IPC codes, with a median of 2 IPC codes per patent (mean 2.6). As shown in Table~\ref{tab:ipc_dist}, the distribution across IPC sections is relatively balanced, ranging from 11.4\% (Section E, Fixed Constructions) to 15.2\% (Sections G and H), with Section D (Textiles; Paper) being the only underrepresented category at 2.6\%. This imbalance reflects the natural distribution of USPTO filings rather than a sampling artifact. Table~\ref{tab:training_stats} provides detailed corpus statistics including text length distributions.

\paragraph{Input Format and Preprocessing.} 
For training, we use the claims section as the primary input. Claims are concatenated in document order, prefixed with a \texttt{[CLS]} token, tokenized using the BERT-for-Patents tokenizer, and truncated to a maximum sequence length of 512 tokens. The CDG is extracted from the available claim text before its token-level projection. When the graph is projected into the truncated input, only claims represented by at least one retained token and edges whose endpoints are represented in the input contribute to the connectivity mask and claim-level objective. Consequently, claims and dependency edges located entirely beyond the 512-token boundary do not provide structural supervision for that training instance. The mean claim-section length is 5{,}937 characters, with a median of 5{,}101 characters, indicating that truncation affects a non-negligible subset of longer patents. We therefore treat limited claim and edge coverage as a limitation of the current fixed-length encoder rather than assuming that the retained prefix captures the complete dependency structure.

\paragraph{Boundary Cases.}
Patents with fewer than two claims account for 2{,}438 instances, or 1.9\% of the training corpus, and contain no inter-claim dependency edges. An additional 591 multi-claim patents, corresponding to 0.4\% of the corpus, contain no edges identified by the extraction pipeline. These 3{,}029 patents use unconstrained self-attention during training and contribute only to the document-level objective, with $\mathcal{L}_{\mathrm{claim}}$ set to zero. We retain these patents to preserve the original corpus distribution and to avoid restricting training to documents for which the extraction pipeline identifies at least one dependency. The remaining 97.7\% of patents contain at least one extracted CDG edge before sequence-length filtering.

\begin{table}[h]
\centering
\small
\caption{Training corpus statistics.}
\label{tab:training_stats}
\begin{tabular}{lr}
\toprule
\textbf{Statistic} & \textbf{Value} \\
\midrule
Total patents         & 131{,}653 \\
Filing years          & 2013--2017 \\
IPC sections          & 8 (A--H) \\
Unique IPC subclasses & 610 \\
Unique IPC codes      & 36{,}094 \\
IPC codes/patent      & 2.6 (med.\ 2) \\
\midrule
\multicolumn{2}{l}{\textit{Text length (chars, mean/med.)}} \\
\quad Title           & 56 / 51 \\
\quad Abstract        & 679 / 697 \\
\quad Claims          & 5{,}937 / 5{,}101 \\
\bottomrule
\end{tabular}
\end{table}

\begin{table}[h]
\centering
\small
\caption{IPC section distribution.}
\label{tab:ipc_dist}
\begin{tabular}{lrr}
\toprule
\textbf{Section} & \textbf{Count} & \textbf{\%} \\
\midrule
A -- Human Necessities     & 19{,}276 & 14.6 \\
B -- Operations; Transport & 18{,}668 & 14.2 \\
C -- Chemistry; Metallurgy & 18{,}226 & 13.8 \\
D -- Textiles; Paper       &  3{,}427 &  2.6 \\
E -- Fixed Constructions   & 15{,}008 & 11.4 \\
F -- Mech.\ Engineering    & 17{,}063 & 13.0 \\
G -- Physics               & 19{,}992 & 15.2 \\
H -- Electricity           & 19{,}993 & 15.2 \\
\bottomrule
\end{tabular}
\end{table}

\section{Detailed DAPFAM Results}
\label{app:dapfam_detail}
TAC queries (title + abstract + claims) consistently outperform TA queries, confirming that claim-inclusive input strengthens cross-patent matching. The best configuration (TAC $\to$ TAC) achieves 0.505 NDCG@100, while Claims $\to$ Claims yields 0.475, indicating that surrounding context (title, abstract) provides complementary retrieval signals beyond claim topology alone. This 0.030 gap suggests that while claim dependencies capture the core inventive structure, titles and abstracts contribute additional lexical cues that aid cross-patent alignment.

In-domain performance is consistently strong across all configurations (0.497--0.508), demonstrating that PHAGE's learned representations generalize well within the same technological field regardless of the specific input sections used for encoding. Out-of-domain scores remain low (0.060--0.063) across all configurations, reflecting the inherent difficulty of cross-domain patent retrieval where inventions in different technological fields share little vocabulary or structural overlap. The stability of these scores across query--corpus pairings indicates that the bottleneck lies in cross-domain semantic distance rather than input format sensitivity.

Notably, the FullText corpus (including the full specification) does not improve over TAC (0.505 vs. 0.505), suggesting that the additional detail in the specification provides no retrieval benefit beyond what is already captured in the title, abstract, and claims.

\begin{table}[h]
\centering
\small
\caption{PHAGE on DAPFAM across query--corpus configurations. IN/OUT: in-domain and out-of-domain NDCG@100.}
\label{tab:dapfam_detail}
\resizebox{\columnwidth}{!}{%
\begin{tabular}{llccccc}
\toprule
\textbf{Query} & \textbf{Corpus} & \textbf{NDCG} & \textbf{R@100} & \textbf{MRR} & \textbf{IN} & \textbf{OUT} \\
\midrule
TA  & TA       & .495 & .336 & .435 & .497 & .062 \\
TA  & TAC      & .494 & .343 & .429 & .496 & .061 \\
TA  & FullText & .495 & .344 & .431 & .497 & .061 \\
TAC & TA       & .496 & .341 & .437 & .499 & .061 \\
TAC & TAC      & .505 & .348 & .448 & .508 & .063 \\
TAC & FullText & .505 & .348 & .447 & .508 & .063 \\
\midrule
\multicolumn{2}{l}{Claims $\to$ Claims} & .475 & .321 & .418 & .477 & .060 \\
\midrule
\multicolumn{2}{l}{\textbf{Average (6 configs)}} & \textbf{.498} & \textbf{.343} & \textbf{.438} & \textbf{.501} & \textbf{.062} \\
\bottomrule
\end{tabular}}
\end{table}

\end{document}